%% file: main.tex
\pdfoutput=1
\documentclass[lettersize,journal]{IEEEtran}
\usepackage{amsmath,amsfonts}
\usepackage{algorithmic}
\usepackage{array}
\usepackage[caption=false,font=normalsize,labelfont=sf,textfont=sf]{subfig}
\usepackage{textcomp}
\usepackage{stfloats}
\usepackage{url}
\usepackage{verbatim}
\usepackage{graphicx}
\usepackage{cite}
\input{preamble.tex}
\hypersetup{pdftitle={STAR-FER: Adaptive Selective Restoration for Edge-Based Low-Resolution Facial Expression Recognition},pdfauthor={ZhongYang Wu, ZhenFeng Gan, Di Zhao, XiaoJun Zhu, HouDe Liu}}
\renewcommand{\abstractname}{Abstract}
\renewcommand{\IEEEkeywordsname}{Index Terms}
\renewcommand{\refname}{References}
\begin{document}
\title{STAR-FER: Adaptive Selective Restoration for Edge-Based Low-Resolution Facial Expression Recognition}
\author{ZhongYang~Wu, ZhenFeng~Gan, Di~Zhao, XiaoJun~Zhu, and HouDe~Liu%
\thanks{ZhongYang Wu and ZhenFeng Gan contributed equally to this work.}%
\thanks{Corresponding author: HouDe Liu.}%
\thanks{The authors are with Tsinghua Shenzhen International Graduate School, Shenzhen 518055, China (e-mail: liu.hd@sz.tsinghua.edu.cn).}}
\bstctlcite{TAC:reference-style}
\maketitle
\begin{abstract}
\input{sections/0_abstract.tex}
\end{abstract}
\input{sections/1_introduction.tex}
\input{sections/2_related_work.tex}
\input{sections/3_method.tex}
\input{sections/4_experiment.tex}
\input{sections/5_conclusion.tex}
\appendices
\input{sections/7_appendix.tex}
\FloatBarrier
\input{sections/6_references.tex}
\end{document}

%% file: preamble.tex
\usepackage{amssymb}
\usepackage{booktabs,multirow}
\usepackage{placeins}
\usepackage[hidelinks]{hyperref}

%% file: sections/0_abstract.tex
Variable interaction distances continuously degrade facial resolution in edge-based affective interaction, making low-resolution facial expression recognition (FER) difficult under limited computation. Full-face image or feature restoration incurs substantial overhead, whereas lightweight recognition pipelines may miss local cues that remain discriminative after degradation. We propose STAR-FER, a scale-and-task-adaptive selective restoration framework that balances recognition performance and resource constraints. A lightweight image--scale conditioned Gate jointly predicts a pixel-wise spatial ranking and a continuous restoration budget, then applies exact-support projection to obtain an input-dependent free-form mask. The mask-conditioned restorer retains complete facial context while concentrating sparse restoration and residual insertion on the selected support. Across seven degradation scales from $\times2$ to $\times8$ on the complete RAF-DB test set, STAR-FER achieves 80.92\% mean accuracy and 71.64\% Macro-F1, improving the low-resolution ResNet-18 baseline by 2.25 and 2.71 percentage points, respectively. It surpasses SR-FER by 1.04 accuracy points while requiring 8.84 GFLOPs, a 54.29\% reduction, with 28.01\% mean mask coverage, and maintains effective recognition at unseen non-integer scales. These results show that spatial-ranking and continuous-budget driven selective restoration reduces spatial modification and computation while maintaining low-resolution FER performance, offering an efficient path to resource-constrained affective interaction.

%% file: sections/1_introduction.tex
\section{Introduction}

\IEEEPARstart{F}{acial} expression recognition (FER) provides an essential perceptual cue for affective human--robot interaction by capturing fine-grained appearance changes around the brows, eyes, and mouth\cite{liu2017hri}. Robots and embedded vision devices operate under limited computation and memory, while the interaction distance can vary continuously. Increasing distance reduces the effective facial pixel scale, producing progressively stronger resolution degradation and erasing subtle expression textures. Upsampling a low-resolution face to a fixed input size changes its dimensions but cannot recover discriminative information that has already been lost. Resource constraints and dynamic interaction distances therefore form a practical deployment bottleneck for low-resolution FER.

Resource-aware FER research mainly reduces inference cost through compact architectures, efficient recognition models, and knowledge distillation\cite{aly2023onlinefer,ma2022distillation}. These methods predict expressions directly from low-resolution inputs and can reduce model size or computation, but they do not explicitly recover local cues lost during degradation. As facial resolution decreases, discriminative changes around the brows, eyes, and mouth are increasingly obscured by blur and noise, limiting the performance attainable by a lightweight classifier alone.

A second line of work addresses resolution degradation through multi-resolution representations, uncertainty modeling, filter learning, and consistency constraints\cite{massoli2021mafer,lo2024uncertainty,yan2020ifsl,lee2024hsac}, or introduces restoration in the image and feature domains. FSR-FER reconstructs super-resolved features from multi-scale low-resolution inputs\cite{nan2022fsrfer}, whereas SR-FER jointly learns image restoration and expression recognition with attention and prediction consistency\cite{xie2024jointfer}. Representative restoration-assisted methods reconstruct a complete image or feature representation, so regions with little expression-discriminative value still participate in processing. Attention and multi-scale representations can emphasize information already present in the input, but they do not directly measure the classification benefit of restoring a candidate region and therefore cannot use that benefit to select the required restoration support.

Existing approaches consequently struggle to meet three requirements together: adaptation to continuously varying degradation scales, restoration restricted to regions that benefit expression recognition, and accurate inference under edge-computing constraints. Without a mechanism that determines restoration support from both degradation severity and task demand, scale adaptation, spatial efficiency, and recognition accuracy remain separately addressed, hindering deployment in dynamic affective interaction.

To close this gap, we propose STAR-FER, a scale-and-task-adaptive selective restoration framework. Its central idea is to determine pixel restoration priority and restoration amount jointly from facial content and degradation scale, concentrating limited restoration computation on discriminative local regions. Our main contributions are as follows:
\begin{enumerate}
 \item We introduce STAR-FER for edge-based affective interaction. It jointly models scale-conditioned region selection, lightweight selective restoration, and expression recognition, replacing fixed full-face processing with local restoration to reduce restoration computation and spatial modification while retaining complete facial context.
 \item We introduce a lightweight image--scale conditioned Gate that learns a free-form spatial ranking under task supervision and regresses a continuous restoration budget from $5\%$ to $35\%$. Exact Top-$k$ projection combines them into an area-controlled, input-dependent restoration mask.
 \item We develop a single-model continuous-scale selective restoration mechanism whose spatial support and restoration amount vary with facial content and degradation scale. Across degradation factors from $\times2$ to $\times8$ on RAF-DB, STAR-FER achieves 80.92\% mean accuracy and 71.64\% Macro-F1 with 8.84 GFLOPs per forward pass.
\end{enumerate}

%% file: sections/2_related_work.tex
\section{Related Work}

STAR-FER draws on two closely related areas: restoration-assisted recognition of low-resolution expressions and spatial allocation of restoration according to the input and task objective.

\subsection{Restoration-Assisted Low-Resolution Facial Expression Recognition}

Super-resolution (SR) provides a foundation for recovering details in general images and faces. The field has progressed from early convolutional reconstruction to deep residual and perceptual restoration models\cite{dong2014srcnn,lim2017edsr,wang2018esrgan}, while blind SR models compound degradations beyond a fixed downsampling kernel\cite{zhang2021bsrgan,wang2021realesrgan}. AdaFM adapts a restoration network to continuously varying restoration levels\cite{he2019adafm}. These methods improve detail reconstruction under changing degradation, but their pixel, perceptual, or adversarial objectives primarily measure image fidelity rather than the value of restored content for expression classification.

Restoration has also been incorporated into low-resolution FER. FSR-FER reconstructs super-resolved features from multi-scale low-resolution inputs\cite{nan2022fsrfer}. SR-FER jointly learns image restoration and expression recognition, transferring task supervision through attention and prediction consistency\cite{xie2024jointfer}. Such methods strengthen degraded representations with reconstructed content, but representative designs still restore a complete facial image or feature representation. Their spatial processing extent is not selected according to the benefit of a region to the current expression decision. Even under joint training, fixed restoration support cannot adapt its spatial extent to the degradation scale and facial instance, so regions with little discriminative contribution still incur computation. STAR-FER instead treats spatial allocation as part of restoration-assisted FER and uses task benefit to determine where restoration modifications are needed.

\subsection{Task-Driven Adaptive Region Selection for Restoration}

Adaptive region-selection methods concentrate visual computation at input-dependent locations. Recurrent visual attention updates subsequent observation locations from previously acquired local information\cite{mnih2014recurrent}. Spatial Transformer Networks predict position, scale, and orientation, then use differentiable resampling to align task-relevant content\cite{jaderberg2015stn}. Spatially sparse dynamic convolutions allocate feature computation at selected locations\cite{verelst2020dynamicconv}. Input-conditioned masks and contour models further represent continuous region geometry using masks, polar distances, or deformable curves\cite{tian2020condinst,xie2020polarmask,peng2020deepsnake}. These approaches support input-dependent feature observation, alignment, sparse computation, or shape representation.

Within image restoration, ClassSR assigns different capacities to image patches according to reconstruction difficulty\cite{kong2021classsr}, whereas SMSR jointly learns spatial and channel masks to identify important reconstruction locations and skip redundant computation\cite{wang2021smsr}. Both demonstrate that restoration computation can be allocated according to image content, but their routing decisions are derived from reconstruction difficulty or image-restoration objectives. STAR-FER learns image-conditioned spatial ranking from classification feedback on restored images and uses training-set task curves to supervise a continuous restoration budget. By combining facial content with degradation scale, the Gate directly predicts restoration locations and coverage, allocating restoration support according to the needs of expression recognition.

%% file: sections/3_method.tex
\section{Method}
\label{sec:method}

\begin{figure*}[t]
 \centering
 \includegraphics[width=\textwidth]{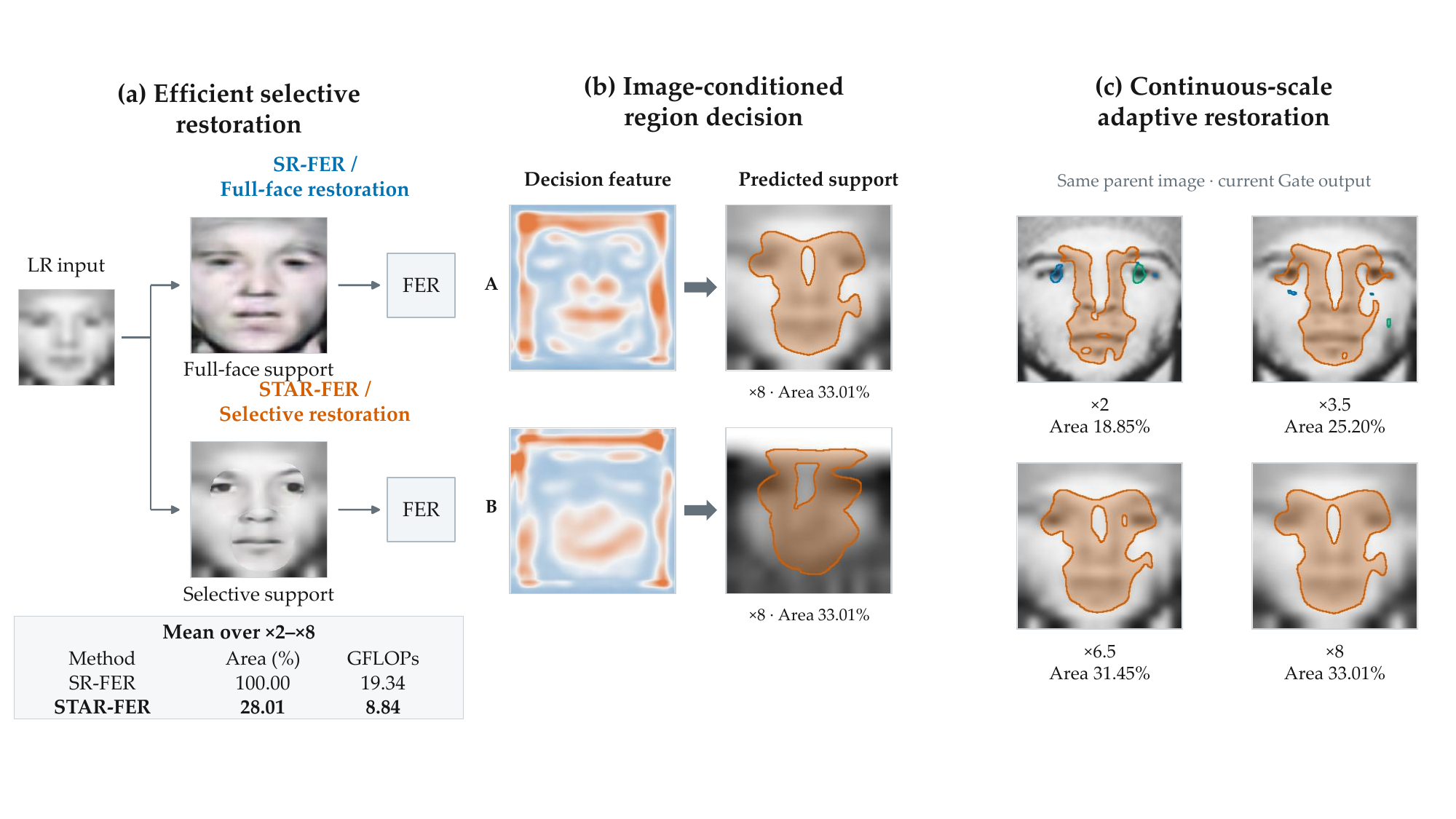}
 \caption{Selective restoration and region decisions in STAR-FER: (a) full-face and selective restoration, (b) Gate decision features and predicted supports for two parent images at the same scale, and (c) predicted supports for one parent image under different degradation scales.}
 \label{fig:lr-fer-motivation}
\end{figure*}

\subsection{Overall Framework}
\label{sec:method-overview}

\begin{figure*}[t]
 \centering
 \includegraphics[width=\textwidth]{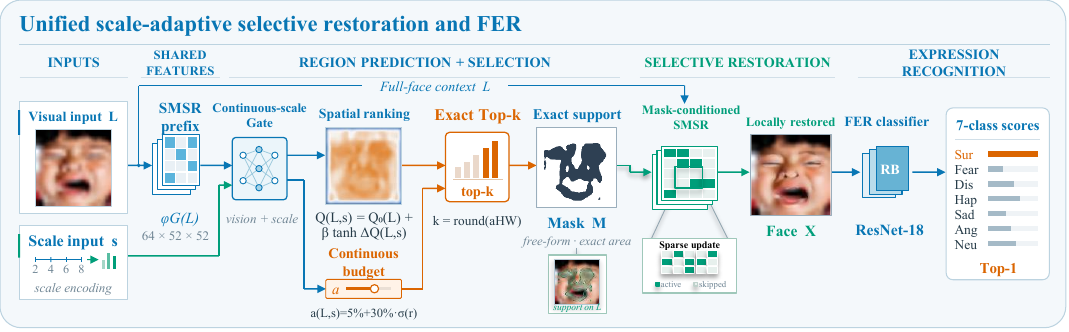}
 \caption{Overview of STAR-FER. The lightweight continuous-scale Gate extracts visual information from a frozen SMSR prefix and jointly predicts a spatial ranking field and an image--scale conditioned restoration budget. Exact Top-$k$ projection converts them into a free-form mask. Mask-conditioned SMSR performs sparse restoration on the selected support, and the locally blended image is passed to a ResNet-18 classifier. Network features and sparse activations are schematic. The reported $28.01\%$ and $8.84$ GFLOPs denote mean coverage across $\times2$--$\times8$ and forward computation under the protocol used in this work, respectively.}
 \label{fig:starfer-overview}
\end{figure*}

Fig.~\ref{fig:lr-fer-motivation} summarizes the three motivations behind STAR-FER: selective restoration, image-conditioned region decisions, and continuous-scale adaptation. As shown in Fig.~\ref{fig:starfer-overview}, the framework comprises a scale-conditioned Gate, a mask-conditioned SMSR restorer, and an FER classifier. Given a low-resolution face $L\in[0,1]^{3\times H\times W}$ and a continuous degradation scale $s$, the Gate $P_\phi$ predicts a spatial ranking field $Q$ and a restoration-area ratio $a$. Exact-budget projection then produces the mask:
\begin{equation}
 (Q,a)=P_\phi(L,s),\qquad
 \widehat M=\operatorname{TopK}\!\left(Q,\operatorname{round}(aHW)\right).
 \label{eq:region-prior}
\end{equation}
The restorer receives the complete low-resolution face together with $\widehat M$ and writes restoration residuals only within the selected support. A seven-class ResNet-18 recognizes the blended image:
\begin{equation}
 \begin{aligned}
 S&=G_\theta(L,\widehat M),\\
 X&=L+\widehat M\odot(S-L),\\
 \widehat y&=\arg\max_k F_\psi(X)_k.
 \end{aligned}
 \label{eq:deployment-forward}
\end{equation}
This unified forward path determines restoration location, restoration amount, and expression class, allowing selective restoration to vary with both the current face and its degradation level. Network dimensions, mask construction, and stage-wise training settings are summarized in Appendix~\ref{app:region-parameters}.

\subsection{Scale-Conditioned Gate: Spatial Ranking and Continuous Budget}
\label{sec:scale-prior}

\begin{figure}[tp]
 \centering
 \includegraphics[width=\columnwidth]{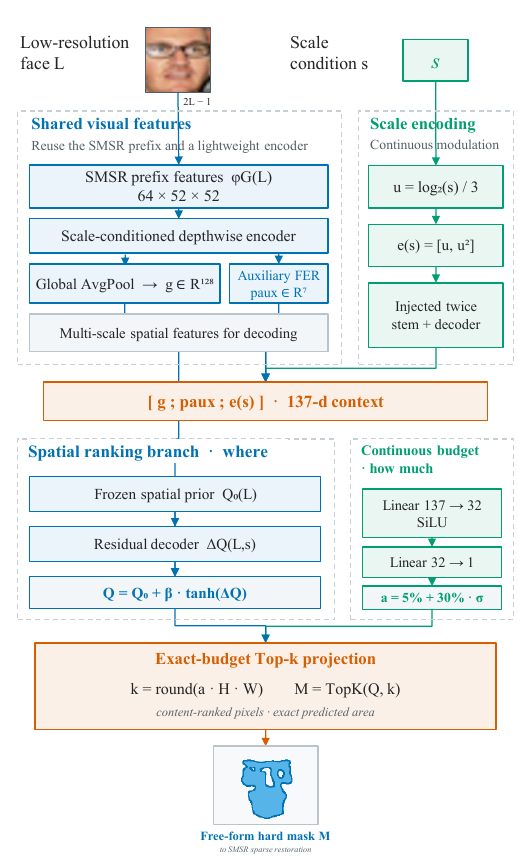}
 \caption{Scale-conditioned Gate. The frozen prefix and source spatial branch provide base features and a source ranking field. The lightweight visual--scale encoder predicts a spatial correction, the continuous budget head estimates restoration area, and exact Top-$k$ generates the final mask.}
 \label{fig:starfer-gate}
\end{figure}

The scale-conditioned Gate must determine both where restoration should occur and how much of the face should be restored. It therefore constructs an image-conditioned spatial ranking from shared visual features and predicts a restoration budget from continuous scale information. Its architecture is shown in Fig.~\ref{fig:starfer-gate}.

\subsubsection{Prefix Features and Visual--Scale Encoding}
The input $L$ is bicubically resized to $52\times52$ and processed by a frozen prefix inherited from the restoration branch:
\begin{equation}
 Z_p=E_{\mathrm{prefix}}(\operatorname{Resize}_{52}(L)),
 \qquad Z_p\in\mathbb R^{64\times52\times52}.
 \label{eq:prefix-feature}
\end{equation}
For video input, the scale condition is computed from the native face-box short side $d_t$ in pixels relative to a 100-pixel reference as $s_t=\operatorname{clip}(100/d_t,2,8)$. The continuous scale is encoded as
\begin{equation}
 u=\frac{\log_2s}{3},\qquad e(s)=[u,u^2],
 \label{eq:scale-code}
\end{equation}
and injected early into the visual branch:
\begin{equation}
 A_0=\operatorname{Stem}(Z_p)+W_{\mathrm{early}}e(s).
 \label{eq:early-scale-modulation}
\end{equation}
Here, $\operatorname{Stem}$ projects the prefix feature to 48 channels, and the scale branch projects $e(s)$ and broadcasts it over the spatial grid. A lightweight encoder built from depthwise convolutions, pointwise convolutions, GroupNorm, and SiLU produces four feature levels, denoted $A$, $B$, $C$, and $D$. Global average pooling of $D$ yields $z\in\mathbb R^{128}$, while an auxiliary head produces seven logits $\ell_{\mathrm{aux}}$. The decoder context is
\begin{equation}
 c=\big[z,\operatorname{sg}(\operatorname{softmax}(\ell_{\mathrm{aux}}/2)),e(s)\big]
 \in\mathbb R^{137},
 \label{eq:gate-context}
\end{equation}
where $\operatorname{sg}$ denotes stop-gradient. The auxiliary response provides semantic conditioning for the Gate, whereas the final expression prediction is produced independently by $F_\psi$. This encoder integrates facial content and continuous degradation scale before spatial support is predicted.

\subsubsection{Image-Conditioned Spatial Ranking}
The spatial branch encodes $Z_p$ together with two-dimensional coordinates to produce a standardized source ranking field $Q_0$. A lightweight decoder fuses the multi-level features $A$, $B$, and $C$ with the prefix feature, $Q_0$, and coordinates, and is modulated by the global context $c$. It predicts a single-channel spatial correction, giving
\begin{equation}
 Q=Q_0+\beta\tanh\!\left(\Delta Q_\phi(Z_p,e(s),c)\right),
 \qquad \beta=0.25.
 \label{eq:spatial-ranking}
\end{equation}
$\Delta Q_\phi$ is produced on a $52\times52$ grid and upsampled to the input resolution. The value of $Q$ determines the relative order in which pixels enter the restoration support. The source field supplies stable spatial structure, while the image-conditioned correction gives different faces distinct rankings at the same scale. This branch thus turns a shared spatial prior into input-dependent restoration locations.

\subsubsection{Continuous Budget and Exact Support}
The restoration amount is predicted from the same context $c$:
\begin{equation}
 a(L,s)=0.05+0.30\,\sigma\!\left(
 W_2\operatorname{SiLU}(W_1c+b_1)+b_2\right),
 \label{eq:continuous-budget}
\end{equation}
where the area head has a $137\rightarrow32\rightarrow1$ architecture and outputs a continuous area ratio between $5\%$ and $35\%$. Let $P=HW$ and $k=\operatorname{round}(aP)$. Selecting the first $k$ pixels after sorting $Q$ in descending order gives
\begin{equation}
 \widehat M(p)=\mathbf1\!\left\{\operatorname{rank}_{\downarrow}(Q(p))<k\right\},
 \qquad \frac1P\sum_p\widehat M(p)=\frac{k}{P}.
 \label{eq:exact-topk}
\end{equation}
This projection converts a continuous budget into area-controlled free-form support. It allows the Gate to learn pixel ranking and restoration amount separately and combine them in one forward pass.

\subsubsection{Differentiable Spatial Learning at a Fixed Budget}
The spatial ranking stage uses a fixed target area $a_0=0.25$ and connects the restoration task to the ranking field through a differentiable mask:
\begin{equation}
 M_p=\sigma\!\left(\kappa[Q_p-t]\right),
 \qquad \frac1P\sum_pM_p=a_0,
 \label{eq:soft-mask}
\end{equation}
where the threshold $t$ is solved for each image by bisection. Accounting for the dependence of $t$ on $Q$, let $g_p=\partial\mathcal L/\partial M_p$ and $w_p=\kappa M_p(1-M_p)$. The backward gradient is
\begin{equation}
 \frac{\partial\mathcal L}{\partial Q_p}
 =w_p\left(g_p-\frac{\sum_qw_qg_q}{\sum_qw_q}\right).
 \label{eq:soft-mask-gradient}
\end{equation}
During training, $\kappa$ increases from 4 to 20 so that the ranking field learns task-related spatial order at a fixed coverage. A task-curve teacher subsequently supervises the continuous budget. This division lets task signals determine both support location and area while retaining exact-budget inference.

\subsection{Lightweight Mask-Conditioned Selective Restoration}
\label{sec:selective-restoration}

Selective restoration must preserve complete facial context while controlling the spatial extent of pixel modification and internal computation. We use SMSR~\cite{wang2021smsr} as the restoration backbone. The Gate mask is mapped to the internal $52\times52$ grid and works with channel selection to modulate eight sparse mixing modules. The complete low-resolution face supplies restoration content throughout the network, and the mask acts as an external spatial condition during the restoration forward pass.

After the restorer produces $S=G_\theta(L,\widehat M)$, its output is locally blended according to~\eqref{eq:deployment-forward}. Pixels outside the support retain their values from $L$, while pixels inside receive the predicted restoration residual. A ResNet-18~\cite{he2016resnet} then performs seven-class FER on the blended image. This module turns Gate support into actual restoration computation and local image modification, connecting region selection to the FER objective.

\subsection{Stage-Wise Multi-Scale Learning}
\label{sec:multiscale-training}

STAR-FER first establishes restoration and recognition ability, then learns image-conditioned spatial ranking, and finally fits the continuous restoration budget. Each stage uses degradation factors from $\times2$ to $\times8$, with stage-specific update sets and supervision objectives.

\subsubsection{Restoration and Recognition Calibration}
SMSR first learns basic restoration through multi-scale pixel supervision and channel-sparsity regularization, followed by joint adaptation with the FER classifier on mask-conditioned inputs. A high-resolution teacher then calibrates the deployed restorer and classifier with
\begin{equation}
 \mathcal L_{\mathrm{calib}}
 =\mathcal L_{\mathrm{CE}}
 +0.5\mathcal L_{\mathrm{KD}}^{T=2}
 +\mathcal L_{\mathrm{HRfeat}}
 +0.05\mathcal L_{\mathrm{masked\,pix}}.
 \label{eq:calibration-loss}
\end{equation}
$\mathcal L_{\mathrm{KD}}^{T=2}$ aligns class distributions at temperature 2, $\mathcal L_{\mathrm{HRfeat}}$ aligns normalized features from the final two FER stages, and $\mathcal L_{\mathrm{masked\,pix}}$ supervises restoration pixels within the selected support. This stage adapts $G_\theta$ and $F_\psi$ to locally blended images and supplies stable task gradients for spatial Gate learning.

\subsubsection{Spatial Gate Learning and Lightweight Distillation}
At a fixed $25\%$ budget, the spatial Gate learns pixel ranking through the classification response to the restored image. Its task supervision is
\begin{equation}
 \begin{aligned}
 \mathcal L_{\mathrm{spatial\,task}}
 ={}&\mathcal L_{\mathrm{CE}}(F_\psi(X),y)
 +0.5\mathcal L_{\mathrm{KD}}^{T=2}
       (F_\psi(X),T_{\mathrm{HR}}(H))\\
 &+\mathcal L_{\mathrm{HRfeat}}.
 \end{aligned}
 \label{eq:spatial-task-loss}
\end{equation}
The auxiliary classifier is supervised by smoothed cross-entropy and the teacher distribution:
\begin{equation}
 \begin{aligned}
 \mathcal L_{\mathrm{aux}}
 ={}&\operatorname{CE}_{\mathrm{smooth}}(\ell_{\mathrm{aux}},y)\\
 &+0.5\mathcal L_{\mathrm{KD}}^{T=2}
       (\ell_{\mathrm{aux}},T_{\mathrm{HR}}(H)).
 \end{aligned}
 \label{eq:auxiliary-loss}
\end{equation}
The lightweight Gate further distills spatial and semantic information from a heavier semantic Gate:
\begin{equation}
 \mathcal L_{\mathrm{distill}}
 =\mathcal L_{\mathrm{Dice}}
 +0.25\mathcal L_{\mathrm{field\,MSE}}
 +0.25\mathcal L_{\mathrm{semantic\,KD}}.
 \label{eq:gate-distillation}
\end{equation}
The Dice term compares masks for the same image and scale, the field loss compares standardized ranking fields, and semantic KD aligns auxiliary responses. The weights of the restorer, FER classifier, source spatial branch, and heavy Gate teacher are frozen, while gradients with respect to mask inputs are retained through the restorer and classifier. This stage converts restoration and recognition feedback into image-conditioned spatial rankings predicted directly by the lightweight Gate.

\subsubsection{Continuous Budget Learning from Task Curves}
After spatial learning, the spatial Gate, restorer, and classifier are fixed and only the continuous budget head is trained. For each training image $i$ and scale, budgets in $\mathcal A=\{0,0.05,\ldots,0.35\}$ are evaluated on the learned ranking field using
\begin{equation}
 J_i(a)=\operatorname{CE}(F_\psi(X_i(a)),y_i)
 +0.5\mathcal L_{\mathrm{KD}}^{T=2}
 +\mathcal L_{\mathrm{HRfeat}}(X_i(a)).
 \label{eq:budget-task-curve}
\end{equation}
The teacher selects the minimum-score budget over the nonzero range from $0.05$ to $0.35$. If the discrete optimum $a_j$ is internal, its neighboring scores $J_-$, $J_0$, and $J_+$ refine the target by a three-point parabola with step $h=0.05$:
\begin{equation}
 a_i^*=\operatorname{clip}_{[.05,.35]}\!\left[
 a_j+h\operatorname{clip}_{[-1,1]}
 \frac{0.5(J_--J_+)}{J_--2J_0+J_+}\right].
 \label{eq:budget-teacher}
\end{equation}
The area head fits $a_i^*$ with
\begin{equation}
 \mathcal L_{\mathrm{budget}}
 =5\mathcal L_{\mathrm{MSE}}
 +0.1\mathcal L_{\mathrm{rank}}
 +5\mathcal L_{\mathrm{moment}}.
 \label{eq:budget-loss}
\end{equation}
The ranking term preserves budget order across images and scales, while the moment term matches the batch mean and standard deviation of predicted and teacher budgets. Task curves provide supervision only on the training set. At deployment, the area head regresses the budget directly from $c$. This final stage compresses image- and degradation-dependent restoration demand into a lightweight forward prediction, enabling continuous-scale selective restoration with one model.

%% file: sections/4_experiment.tex
\section{Experiments}
\label{sec:experiments}

\begin{figure*}[t]
 \centering
 \includegraphics[width=\textwidth]{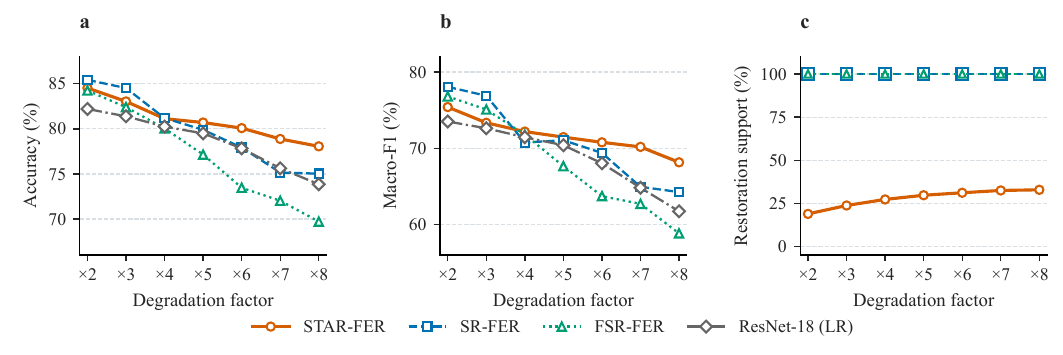}
 \caption{Recognition and base support across seven RAF-DB scales. (a) Accuracy; (b) Macro-F1; (c) Mask coverage.}
 \label{fig:rafdb-performance}
\end{figure*}

\begin{figure*}[t]
 \centering
 \includegraphics[width=\textwidth]{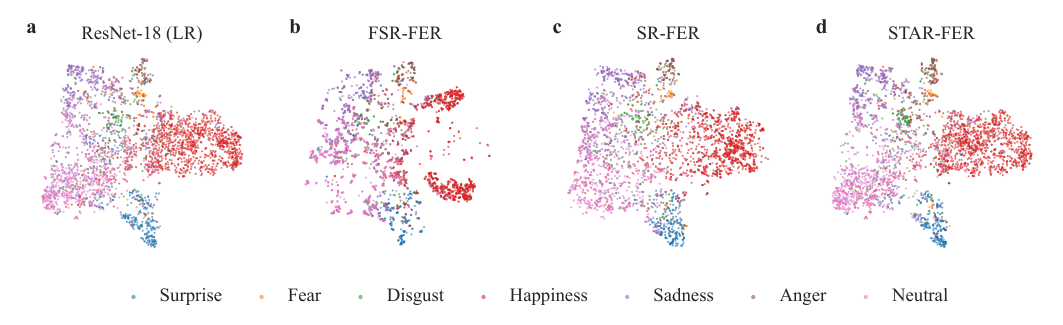}
 \caption{t-SNE visualization of RAF-DB features at $\times8$.}
 \label{fig:feature-distributions}
\end{figure*}

\subsection{Datasets}
\label{sec:experiment-setup}

\noindent\textbf{RAF-DB} RAF-DB is a large-scale in-the-wild facial expression dataset containing approximately 30,000 face images. Its basic and compound expression annotations were obtained through repeated crowdsourcing and reliability filtering\cite{li2017rafdb}. Following the standard basic-expression protocol, we use the seven-class subset with 12,271 training and 3,068 test images.

\noindent\textbf{FERPlus} FERPlus relabels FER2013 using independent votes from ten annotators per image, yielding a label distribution over eight expression categories\cite{barsoum2016ferplus}. The dataset contains 32,298 training and 3,589 test images.

Both datasets are evaluated at seven integer degradation scales from $\times2$ to $\times8$, with equal weighting across scales. FERPlus recognition is evaluated on the cleaned seven-class subset containing 3,136 test images. Data processing, region parameters, and training settings are provided in Appendix~\ref{app:region-parameters}.

\subsection{Recognition Results and Analysis}
\label{sec:experiment-main}

This section examines whether selective restoration improves low-resolution FER and how its effect changes with degradation severity. We compare STAR-FER with ResNet-18 applied directly to low-resolution inputs, the image-restoration method SR-FER\cite{xie2024jointfer}, and the feature-restoration method FSR-FER\cite{nan2022fsrfer} from $\times2$ to $\times8$. Each scale uses all 3,068 RAF-DB test images, and the cross-scale mean equally weights the seven scales. STAR-FER uses BF16 for G and the Gate and FP32 for the FER classifier during inference.

\begin{table}[!htb]
 \centering
 \caption{Mean recognition results across RAF-DB scales.}
 \label{tab:overall-recognition}
 \begin{tabular*}{\columnwidth}{@{\extracolsep{\fill}}lrr@{}}
 \toprule
 Method & Accuracy (\%) & Macro-F1 (\%)\\
 \midrule
 ResNet-18 (LR) & 78.66 & 68.94\\
 STAR-FER  & \textbf{80.92} & \textbf{71.64}\\
 SR-FER & 79.88 & 70.76\\
 FSR-FER & 77.01 & 68.09\\
 \bottomrule
 \end{tabular*}
\end{table}

As shown in Table~\ref{tab:overall-recognition}, STAR-FER achieves mean accuracy of 80.92\% and Macro-F1 of 71.64\%. It improves over the low-resolution baseline by 2.25 and 2.71 percentage points, SR-FER by 1.04 and 0.88 points, and FSR-FER by 3.91 and 3.56 points, respectively. With local restoration support, STAR-FER also achieves higher mean recognition scores than full-image and full-feature restoration methods, showing that low-resolution FER can match or surpass prior methods without restoring every spatial location to the same extent.

Fig.~\ref{fig:rafdb-performance}(a)(b) further shows how recognition changes with degradation scale. As degradation becomes more severe, less discriminative information remains and all four methods decline; however, STAR-FER exhibits the most stable trend. From $\times2$ to $\times8$, it loses 6.47 accuracy points and 7.24 Macro-F1 points. The corresponding declines are 8.34 and 11.77 points for the low-resolution baseline, 10.40 and 13.79 points for SR-FER, and 14.53 and 17.95 points for FSR-FER. Although SR-FER leads STAR-FER by 0.90 accuracy points at $\times2$, STAR-FER reaches 78.06\% at $\times8$ and exceeds SR-FER and FSR-FER by 3.03 and 8.34 points. The smaller cross-scale decline and sustained accuracy under severe degradation support the role of scale-conditioned region selection in retaining stable expression evidence as facial details disappear.

Fig.~\ref{fig:rafdb-performance}(c) relates recognition performance to restoration extent. STAR-FER uses 28.01\% mean mask coverage, whereas SR-FER and FSR-FER process the complete image and feature space, respectively. As degradation increases from $\times4$ to $\times8$, STAR-FER expands its coverage from 27.27\% to 32.85\%, indicating that the Gate allocates more restoration support when visual cues become weaker while retaining compact support under milder degradation. Together with the recognition gains in Table~\ref{tab:overall-recognition}, this result shows that scale-dependent allocation removes most spatial modifications while preserving local information useful for FER.

To examine feature organization under severe degradation, Fig.~\ref{fig:feature-distributions} projects features from the same 3,068 RAF-DB test images at $\times8$ into two dimensions. Happiness forms a relatively compact cluster, and surprise also exhibits distinct grouping, whereas neutral and sadness still overlap considerably. Happiness samples in FSR-FER split into several subclusters, while SR-FER and STAR-FER exhibit more continuous within-class structures. This pattern is consistent with their stronger recognition under severe degradation and illustrates how restoration changes within-class organization. The remaining overlap between neutral and sadness also identifies classes that the current representations still confuse.

\begin{table}[!t]
 \centering
 \caption{Recognition performance and Gate coverage on FERPlus. F1 denotes Macro-F1.}
 \label{tab:ferplus-accuracy}
 \setlength{\tabcolsep}{2pt}
 \begin{tabular*}{\columnwidth}{@{\extracolsep{\fill}}crrrrr@{}}
 \toprule
 & \multicolumn{3}{c}{STAR-FER} & \multicolumn{2}{c}{ResNet-18 (LR)}\\
 \cmidrule(lr){2-4}\cmidrule(lr){5-6}
 Scale & Acc (\%) & F1 (\%) & Coverage (\%) & Acc (\%) & F1 (\%)\\
 \midrule
 $\times2$ & 85.71 & 75.67 & 17.23 & 85.68 & 75.72\\
 $\times3$ & 85.17 & 74.92 & 23.20 & 85.14 & 74.89\\
 $\times4$ & 84.34 & 74.45 & 28.87 & 84.02 & 74.02\\
 $\times5$ & 83.07 & 72.73 & 33.64 & 82.53 & 72.65\\
 $\times6$ & 82.68 & 72.25 & 34.77 & 80.80 & 68.62\\
 $\times7$ & 81.98 & 72.27 & 34.84 & 77.71 & 67.03\\
 $\times8$ & 79.91 & 69.81 & 34.88 & 72.54 & 58.13\\
 \midrule
 Mean & \textbf{83.27} & 73.16 & 29.63 & 81.20 & 70.15\\
 \bottomrule
 \end{tabular*}
\end{table}

Table~\ref{tab:ferplus-accuracy} shows that the proposed method achieves mean accuracy of 83.27\% and Macro-F1 of 73.16\% across the seven scales, with mean coverage of 29.63\%.

\subsection{Ablation Experiments and Mechanism Analysis}
\label{sec:experiment-ablation}

\begin{figure}[!t]
 \centering
 \includegraphics[width=\columnwidth]{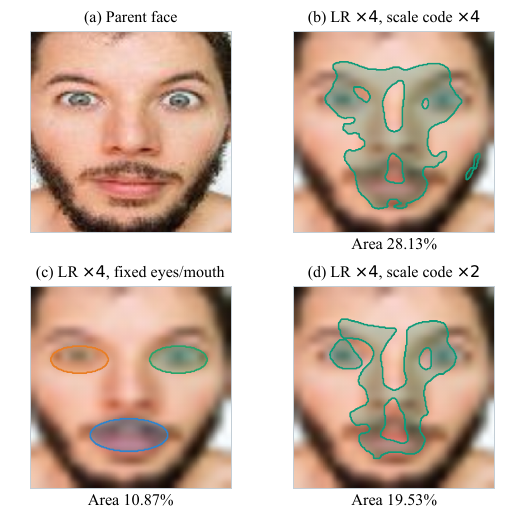}
 \caption{Illustration of fixed eye-and-mouth support and adaptive selection. (a) Parent face. (b) Adaptive mask for a $\times4$ input with its corresponding scale condition. (c) Fixed eye-and-mouth mask on the same $\times4$ input. (d) Adaptive mask with the input kept at $\times4$ and the scale code fixed to $\times2$.}
 \label{fig:fixed-eye-control}
\end{figure}

\subsubsection{Analysis of Scale-Encoding Benefits}
To examine how scale encoding affects restoration budgets and recognition, we keep the degraded images at each scale unchanged and replace the normal Gate scale code with a constant $\times2$ code. Table~\ref{tab:scale-code-ablation} reports the comparison on the complete RAF-DB test set at seven degradation scales.

\begin{table}[!htb]
 \centering
 \caption{Normal versus constant $\times2$ scale encoding. The actual input degradation is unchanged.}
 \label{tab:scale-code-ablation}
 \setlength{\tabcolsep}{3pt}
 \begin{tabular*}{\columnwidth}{@{\extracolsep{\fill}}crrrr@{}}
 \toprule
 \multirow{2}{*}{Input scale} & \multicolumn{2}{c}{Accuracy (\%)} & \multicolumn{2}{c}{Coverage (\%)}\\
 \cmidrule(lr){2-3}\cmidrule(lr){4-5}
 & Normal & Fixed $\times2$ & Normal & Fixed $\times2$\\
 \midrule
 $\times2$ & 84.53 & 84.53 & 18.89 & 18.89\\
 $\times3$ & 83.04 & 83.94 & 23.76 & 18.93\\
 $\times4$ & 81.13 & 80.87 & 27.27 & 19.09\\
 $\times5$ & 80.70 & 80.61 & 29.70 & 19.33\\
 $\times6$ & 80.08 & 79.53 & 31.14 & 19.34\\
 $\times7$ & 78.88 & 78.23 & 32.48 & 19.50\\
 $\times8$ & 78.06 & 76.76 & 32.85 & 19.28\\
 \midrule
 Mean & \textbf{80.92} & 80.64 & 28.01 & 19.19\\
 \bottomrule
 \end{tabular*}
\end{table}

Table~\ref{tab:scale-code-ablation} shows that normal encoding increases mean coverage from 18.89\% at $\times2$ to 32.85\% at $\times8$. With the code fixed to $\times2$, coverage stays between 18.89\% and 19.50\%, reducing its seven-scale mean from 28.01\% to 19.19\%. Scale conditioning therefore guides restoration-budget allocation as degradation increases. Fixing the code to a mild degradation level makes the model allocate an amount close to that for $\times2$, even for more blurred inputs. The decrease from 28.13\% to 19.53\% for the same $\times4$ face in Fig.~\ref{fig:fixed-eye-control}(b)(d) illustrates this budget response.

Across the seven scales, mean accuracy is 80.92\% with normal scale encoding and 80.64\% with fixed encoding, a difference of 0.28 percentage points. Normal encoding yields accuracy higher by 0.55, 0.65, and 1.30 percentage points at $\times6$, $\times7$, and $\times8$, respectively. Meanwhile, mean mask coverage differs by 8.82 percentage points, with larger differences in restoration extent under severe degradation. These results indicate that fixed encoding restricts the Gate's ability to adjust restoration extent to degradation severity. Under mild degradation, the input retains substantial discriminative information, so restricting restoration extent has a relatively small effect on recognition. As degradation intensifies, fine-grained expression cues disappear, and limited restoration support becomes less effective at compensating for the loss of discriminative information, leading to a larger performance decline. Scale conditioning instead allocates the restoration budget according to input degradation, providing more extensive local support under severe low-resolution conditions and mitigating its effect on recognition.

We further fix spatial ranking while retaining each image's own predicted budget to examine the source of location differences. Replacing image-conditioned ranking with a fixed learned template increases mean cross-parent mask IoU from 0.5705 to 0.9586. Higher IoU indicates more similar supports across faces. The near-unit IoU after replacement therefore shows that restoration locations become largely shared, with the remaining differences mainly arising from boundary-size changes induced by per-image budgets. The image-conditioned spatial branch is the main source of personalized locations through its adaptive pixel ranking, while scale encoding primarily regulates restoration amount.

\begin{table}[!t]
 
 \begingroup
 \centering
 \caption{RAF-DB accuracy of the adaptive Gate and fixed eye-and-mouth support across seven scales. $\Delta$Acc denotes STAR-FER minus fixed accuracy.}
 \label{tab:fixed-eye-recognition}
 \setlength{\tabcolsep}{3pt}
 \begin{tabular*}{\columnwidth}{@{\extracolsep{\fill}}crrr@{}}
 \toprule
 Scale & \shortstack{STAR-FER\\Acc (\%)} & \shortstack{Fixed\\Acc (\%)} & $\Delta$Acc (pp)\\
 \midrule
 $\times2$ & 84.53 & 84.63 & $-0.10$\\
 $\times3$ & 83.04 & 83.84 & $-0.80$\\
 $\times4$ & 81.13 & 81.02 & $+0.11$\\
 $\times5$ & 80.70 & 80.44 & $+0.26$\\
 $\times6$ & 80.08 & 79.37 & $+0.71$\\
 $\times7$ & 78.88 & 77.77 & $+1.11$\\
 $\times8$ & 78.06 & 75.97 & $+2.09$\\
 \midrule
 Mean & \textbf{80.92} & 80.43 & $\mathbf{+0.48}$\\
 \bottomrule
 \end{tabular*}
 \par\endgroup
\end{table}

\subsubsection{Adaptive Gate versus Fixed Eye-and-Mouth Regions}
We compare fixed anatomical support with adaptive selection using the same restorer and FER classifier.

The fixed support consists of three ellipses covering the eyes and mouth. Evaluation uses all 3,068 RAF-DB test images at seven scales. Table~\ref{tab:fixed-eye-recognition} reports scale-wise recognition results, and Fig.~\ref{fig:fixed-eye-control}(b)(c) illustrates the two supports.

As shown in Table~\ref{tab:fixed-eye-recognition}, STAR-FER achieves mean accuracy of 80.92\%, compared with 80.43\% for fixed eye-and-mouth support, a gain of 0.48 percentage points in this comparison. STAR-FER is lower by 0.10 and 0.80 points at $\times2$ and $\times3$, respectively. It outperforms fixed support from $\times4$ onward, with gains of 0.71, 1.11, and 2.09 points at $\times6$, $\times7$, and $\times8$. Accuracy declines by 8.66 points from $\times2$ to $\times8$ with fixed support, compared with 6.47 points for STAR-FER. Adaptive selection therefore retains recognition performance more effectively under severe degradation.

This pattern indicates that, under mild degradation, the eyes and mouth retain substantial expression cues, and restoring these key locations satisfies most recognition needs. Fixed eye-and-mouth support can therefore match or slightly outperform adaptive selection. As resolution decreases further, local textures and fine-grained expression information disappear, making fixed eye-and-mouth restoration less effective at compensating for the lost discriminative information. STAR-FER instead adjusts restoration locations and coverage according to facial content and degradation severity, allocating support to regions more useful for expression recognition. The adaptive spatial support in Fig.~\ref{fig:fixed-eye-control}(b)(c), together with the increasing recognition advantage under severe degradation in Table~\ref{tab:fixed-eye-recognition}, shows that adaptive region selection mitigates the limitations of fixed eye-and-mouth restoration and better preserves recognition performance at very low resolutions.

\subsubsection{Dense versus Sparse Restoration Computation}
To examine whether selected support translates into computational savings, we fix network weights and masks and compare the dense masked and true SMSR sparse paths of the same model. Table~\ref{tab:sparse-execution-ablation} reports computation including the Gate, restorer, and FER classifier.

\begin{table}[!htb]
 \centering
 \caption{Execution-path comparison under identical weights and masks. Computation is averaged over seven scales for the complete pipeline.}
 \label{tab:sparse-execution-ablation}
 \setlength{\tabcolsep}{3pt}
 \begin{tabular*}{\columnwidth}{@{\extracolsep{\fill}}lrr@{}}
 \toprule
 Execution path & Cost (GFLOPs) & Reduction (\%)\\
 \midrule
 Dense masked & 9.5646 & \textemdash\\
 True SMSR sparse & \textbf{8.8380} & \textbf{7.60}\\
 \bottomrule
 \end{tabular*}
\end{table}

Table~\ref{tab:sparse-execution-ablation} shows that true sparse execution reduces mean computation from 9.5646 to 8.8380 GFLOPs, saving 0.7266 GFLOPs, or 7.60\%. With identical restoration support, the dense masked path still performs convolutions over the complete spatial grid, whereas the sparse path skips part of the computation according to spatial support and channel selection. Reducing restoration area and reducing convolutional operations thus serve distinct roles: the Gate determines local restoration extent, and sparse SMSR execution translates this spatial selection into computational savings. This result supports combining spatial selection with sparse execution in a lightweight restoration backbone to reduce forward computation.

\subsection{Computational Efficiency}
\label{sec:experiment-runtime}

\begin{table}[!htb]
 \centering
 \caption{Model parameters, mean restoration coverage, and computational cost.}
 \label{tab:model-complexity}
 \setlength{\tabcolsep}{2pt}
 \begin{tabular*}{\columnwidth}{@{\extracolsep{\fill}}lrrr@{}}
 \toprule
 Method & Parameters (M) & Coverage (\%) & Cost (GFLOPs)\\
 \midrule
 STAR-FER & 12.885 & \textbf{28.01} & \textbf{8.84}\\
 SR-FER & 14.21 & 100.00 & 19.34\\
 FSR-FER & 35.42 & 100.00 & 19.41\\
 \bottomrule
 \end{tabular*}
\end{table}

Table~\ref{tab:model-complexity} compares model size, mean restoration coverage, and computational cost. STAR-FER contains 12.885 M parameters and performs selective restoration over 28.01\% mean spatial support, requiring 8.84 GFLOPs for one forward pass through the Gate, restorer, and FER classifier. SR-FER and FSR-FER process the complete image or feature space and require 19.34 and 19.407 GFLOPs, respectively. STAR-FER therefore reduces computation by 54.29\% and 54.45\%. Together with the preceding recognition results, these measurements show that STAR-FER substantially reduces restoration extent and computational cost while maintaining recognition performance, making it better suited to affective interaction on edge devices.

\subsection{Continuous-Scale Adaptation Analysis}
\label{sec:continuous-scale-analysis}

\begin{figure*}[!t]
 \centering
 \includegraphics[width=\textwidth]{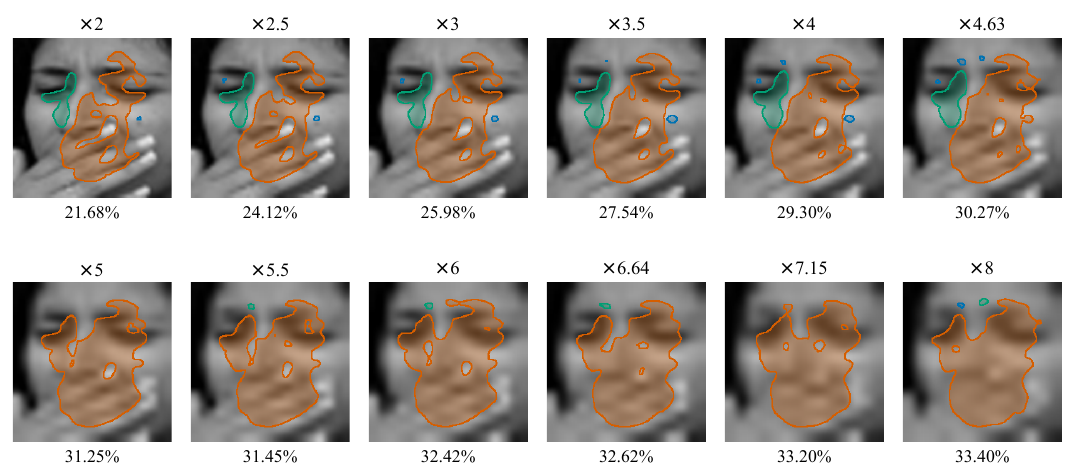}
 \caption{Restoration masks predicted by the current continuous-scale Gate for the same parent image across degradation scales. Orange, green, and blue denote the three largest connected components in descending area; values below each image give union-mask coverage.}
 \label{fig:continuous-scale-regions}
\end{figure*}

Multi-scale training uses only the seven integer factors from $\times2$ to $\times8$. To test direct extrapolation of the continuous scale condition, Table~\ref{tab:continuous-scale-generalization} evaluates the frozen model at three unseen non-integer scales. Fig.~\ref{fig:continuous-scale-regions} further visualizes continuous responses between the integer training scales for the same parent image.

\begin{table}[t]
 \centering
 \caption{RAF-DB recognition and Gate coverage at unseen non-integer scales.}
 \label{tab:continuous-scale-generalization}
 \footnotesize
 \setlength{\tabcolsep}{2.5pt}
 \begin{tabular*}{\columnwidth}{@{\extracolsep{\fill}}crrr@{}}
 \toprule
 Scale & Acc (\%) & F1 (\%) & Coverage (\%) \\
 \midrule
 $\times4.55$ & 80.35 & 71.50 & 28.60 \\
 $\times6.25$ & 79.47 & 70.27 & 31.57 \\
 $\times7.14$ & 78.10 & 69.31 & 32.52 \\
 \bottomrule
 \end{tabular*}
\end{table}

Table~\ref{tab:continuous-scale-generalization} shows that, as the effective factor increases from $\times4.55$ to $\times7.14$, mean Gate coverage rises from 28.60\% to 32.52\% while accuracy decreases smoothly from 80.35\% to 78.10\%. These unseen-scale results indicate that continuous scale conditioning adjusts the restoration budget between adjacent factors rather than merely memorizing the seven training scales.

Fig.~\ref{fig:continuous-scale-regions} shows the regions directly predicted by the current Gate for the same parent image at 12 scales from $\times2$ to $\times8$. Union-mask coverage increases from $21.68\%$ to $33.40\%$, while the number, location, and contour of connected regions vary with scale. This example shows that continuous scale conditioning changes both the restoration budget and the spatial distribution of the free-form support.

\subsection{Scale Budget and Instance Variation Analysis}
\label{sec:intervention-quality}

Beyond mean recognition scores, we measure region-area variation and cross-image mask IoU for the same frozen Gate on 3,068 RAF-DB test images to examine whether it produces personalized regions or a single fixed template.

\begin{table}[t]
 \centering
 \caption{Scale-wise RAF-DB Gate-region statistics.}
 \label{tab:gate-region-stats}
 \setlength{\tabcolsep}{1.5pt}
 \begin{tabular*}{\columnwidth}{@{\extracolsep{\fill}}crrr@{}}
 \toprule
 Scale & Coverage (\%) & Std. (\%) & Cross-IoU\\
 \midrule
 $\times2$ & 18.89 & 1.56 & 0.391\\
 $\times3$ & 23.76 & 1.48 & 0.484\\
 $\times4$ & 27.27 & 1.22 & 0.554\\
 $\times5$ & 29.70 & 0.99 & 0.604\\
 $\times6$ & 31.14 & 0.78 & 0.632\\
 $\times7$ & 32.48 & 0.56 & 0.658\\
 $\times8$ & 32.85 & 0.48 & 0.669\\
 \midrule
 Mean & 28.01 & 1.01 & 0.570\\
 \bottomrule
 \end{tabular*}
\end{table}

Table~\ref{tab:gate-region-stats} shows mean Gate coverage of 28.01\%, increasing with degradation severity. Cross-image IoU ranges from 0.391 to 0.669, indicating that the predicted regions vary across images rather than collapsing to a single fixed template. Its increase at stronger degradation indicates that different images share more restoration support when visual evidence is weak. These results further show that the method learns personalized restoration regions for different facial appearances, adapting its support to expression recognition across faces.

\subsection{Real-World Deployment and Evaluation}
\label{sec:real-world}

\subsubsection{PC-Based Video Evaluation}
To examine restoration responses and recognition outputs under continuous input, we evaluate a recorded video on a PC equipped with an Intel Core i5-12500H processor and an NVIDIA GeForce RTX 3050 Ti Laptop GPU. The three methods use the same video, face detection, and cropping conditions. We compare adjacent-frame prediction switching rates and mean method latency.

\begin{figure}[!t]
 \centering
 \includegraphics[width=\columnwidth]{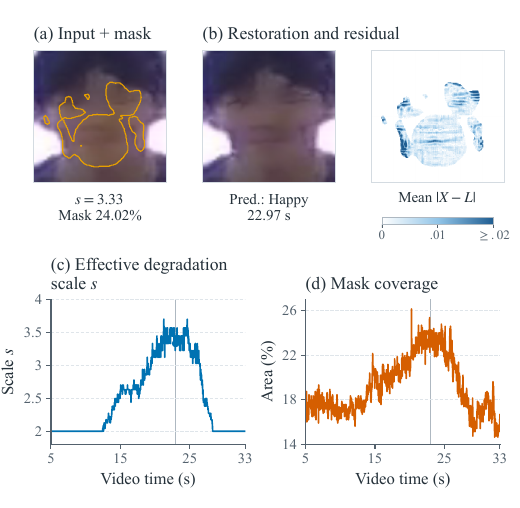}
 \caption{Selective restoration and scale response in the PC deployment experiment. (a) Input with the predicted mask. (b) Locally fused result $X$ and channel-mean $|X-L|$ to its right. (c) Effective degradation scale $s$. (d) Mask coverage. Curves show valid observations from 5 to 33 s; dots mark 22.97 s, with $s=3.33$ and 24.02\% coverage.}
 \label{fig:pc-video-response}
\end{figure}

Fig.~\ref{fig:pc-video-response}(a)(b) shows the input and locally fused result at 22.97 s, with a native face short side of 30 pixels, a scale condition of 3.33, and mask coverage of 24.02\%. The difference map shows that restoration changes are confined to the predicted support, consistent with the local fusion in~\eqref{eq:deployment-forward}. Fig.~\ref{fig:pc-video-response}(c)(d) shows the continuous response from 5 to 33 s of the original video. The effective degradation scale rises from 2 to approximately 3.7 and then returns to 2, while mask coverage generally increases and then decreases, fluctuating between approximately 15\% and 26\%. The plateaus at 2 in the early and late intervals result from the lower bound on the scale condition. As effective degradation increases, the Gate generally expands restoration support; as the facial pixel scale recovers, the support contracts. This response shows that the scale-conditioned Gate adapts restoration budgets and spatial support to changes in the effective facial pixel scale.

\begin{figure}[!htb]
 \centering
 \includegraphics[width=\columnwidth]{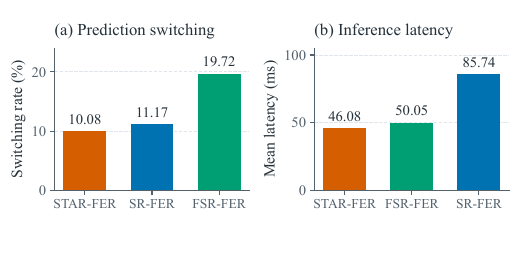}
 \caption{Overall PC deployment statistics. (a) Prediction switching rates over 913 adjacent valid-frame pairs. (b) Mean method latency over 914 detected faces.}
 \label{fig:pc-video-test}
\end{figure}

Fig.~\ref{fig:pc-video-test}(a) shows a prediction switching rate of 10.08\% for STAR-FER, compared with 11.17\% for SR-FER and 19.72\% for FSR-FER, reductions of 1.09 and 9.64 percentage points, respectively. Under the same video, detection, and cropping conditions, STAR-FER switches classes less frequently, producing more consistent predictions, particularly relative to FSR-FER. These observations indicate that selective restoration maintains comparatively stable expression predictions as the facial pixel scale changes continuously.

Fig.~\ref{fig:pc-video-test}(b) reports method latency. STAR-FER requires 46.08 ms on average, reducing latency by 3.97 ms (7.93\%) relative to FSR-FER at 50.05 ms and by 39.66 ms (46.26\%) relative to SR-FER at 85.74 ms. Together with the switching-rate results, this PC-based video evaluation shows that STAR-FER achieves both fewer class switches and lower method latency, supporting the practical value of selective restoration under continuous visual input.

\subsubsection{Edge Deployment and Evaluation}
To examine inference efficiency on a resource-constrained device, we conduct edge deployment tests on an NVIDIA Jetson Orin Nano Super Developer Kit. The robot uses a DECXIN USB Camera with 1080p imaging and a $90^{\circ}$ distortion-free field of view. The software environment comprises Ubuntu 22.04.5 LTS, Jetson Linux 36.4.3, and CUDA 12.6.

We compare ResNet-18 (LR), SR-FER, FSR-FER, and the dense masked and sparse execution paths of STAR-FER. Single-frame inference uses a batch size of 1. Table~\ref{tab:jetson-deployment} reports mean latency, P95 latency, and peak CUDA tensor allocation measured by PyTorch.

\begin{table}[!htb]
 \centering
 \caption{Single-frame inference on Jetson Orin Nano Super with batch size 1. Memory denotes peak CUDA tensor allocation measured by PyTorch.}
 \label{tab:jetson-deployment}
 \setlength{\tabcolsep}{2pt}
 \begin{tabular*}{\columnwidth}{@{\extracolsep{\fill}}lrrr@{}}
 \toprule
 Method & Mean (ms) & P95 (ms) & \shortstack{Peak CUDA\\memory (MB)}\\
 \midrule
 ResNet-18 (LR) & 15.37 & 23.58 & 92.40\\
 SR-FER & 132.06 & 141.51 & 727.54\\
 FSR-FER & 146.21 & 149.54 & 202.75\\
 STAR-FER (Dense) & \textbf{85.46} & \textbf{88.93} & 115.29\\
 STAR-FER (Sparse) & 122.87 & 127.53 & \textbf{115.26}\\
 \bottomrule
 \end{tabular*}
\end{table}

As shown in Table~\ref{tab:jetson-deployment}, ResNet-18 applied directly to low-resolution inputs has the lowest mean latency at 15.37 ms. Among restoration-assisted methods, the dense masked path of STAR-FER takes 85.46 ms on average, reducing latency by 35.29\% relative to SR-FER and 41.55\% relative to FSR-FER. The sparse path takes 122.87 ms, with corresponding reductions of 6.96\% and 15.96\%. Their P95 latencies of 88.93 and 127.53 ms are also below the 141.51 ms of SR-FER and 149.54 ms of FSR-FER. These findings are consistent with the preceding analysis of computational savings from the compact restoration architecture and selective restoration path, showing that the design delivers practical latency reductions over full-face restoration baselines on Jetson. The reductions in both mean and P95 latency extend this advantage from average processing efficiency to frames requiring longer inference times, helping reduce response delays during on-device interaction. Together with the recognition results, these measurements support the feasibility of deploying STAR-FER on resource-constrained affective interaction devices: it recovers local expression cues and maintains recognition performance while reducing the inference burden on the target hardware.

Table~\ref{tab:jetson-deployment} also shows similar peak CUDA tensor allocations for the dense masked and sparse paths, at 115.29 and 115.26 MB, respectively. The sparse path reduces this peak by 84.16\% relative to SR-FER at 727.54 MB and by 43.15\% relative to FSR-FER at 202.75 MB. Together with the latency measurements, these results show that STAR-FER reduces inference time and CUDA tensor memory demand on the evaluated edge platform, while switching to sparse execution has little effect on peak allocation.

Despite its lower theoretical operation count, the sparse path has 37.41 ms higher mean latency than the dense masked path. Module-level measurements show similar times for the Gate, Top-$k$ mask generation, and FER classifier. In contrast, the SMSR restorer and mask blending increase from 43.59 to 80.31 ms, an additional 36.72 ms that accounts for most of the total latency gap. PyTorch operator records further identify mask indexing, convolution-window unfolding, matrix multiplication, and result write-back in the current sparse implementation. Scheduling and data-access overhead from these operations may offset part of the convolutional savings, limiting the hardware speedup obtained from the lower operation count. The dense masked path consequently provides better inference efficiency in the current implementation. The sparse path still requires optimization of data organization and operator execution, which remains a limitation of the method.

\FloatBarrier

%% file: sections/5_conclusion.tex
\section{Conclusion}
We presented STAR-FER, a computationally lightweight FER framework for on-device deployment. Its scale-conditioned Gate jointly predicts an image-dependent spatial ranking and a continuous restoration budget, which exact-support projection converts into a free-form mask. The mask-conditioned restorer performs local sparse restoration with complete facial context, followed by FER classification. Across seven degradation scales on the complete RAF-DB test set, STAR-FER achieves 80.92\% mean accuracy, 71.64\% Macro-F1, and 28.01\% mean coverage with 8.84 GFLOPs per forward pass. Continuous-scale and instance-level region statistics further show that one model can adjust restoration amount with degradation severity and produce different spatial supports for different inputs.

\label{sec:limitations}

\noindent\textbf{Limitations and Future Work:}
The current Gate produces instance-varying restoration regions, but their spatial differences have not yet translated into consistent recognition gains over a strong fixed spatial prior. Jetson measurements further show that the current sparse SMSR path is slower than dense masked execution, so its theoretical savings have not fully translated into hardware speedup. Future work will further strengthen the connection between spatial ranking and classification benefit.

%% file: sections/7_appendix.tex
\section{Implementation Details and Data Protocol}
\label{app:region-parameters}

\subsection{Gate Dimensions and Spatial-Field Processing}

After resizing the input to $52\times52$, the frozen prefix produces a $64\times52\times52$ feature map. The four levels of the lightweight encoder have dimensions $48\times52\times52$, $64\times26\times26$, $96\times13\times13$, and $128\times7\times7$. The global feature, seven-dimensional auxiliary class distribution, and two-dimensional scale code form the 137-dimensional context. The spatial decoder projects each of the first three feature levels and the prefix feature to 24 channels and concatenates them with the single-channel source field and two coordinate channels. A $137\rightarrow48$ context projection modulates the decoder. The area head has a $137\rightarrow32\rightarrow1$ architecture with 4,449 trainable parameters.

Spatial fields are standardized independently for each image. For a field $Q$ with spatial mean $\mu_Q$ and standard deviation $\sigma_Q$,
\begin{equation}
 \operatorname{Norm}(Q)=
 \frac{Q-\mu_Q}{\max(\sigma_Q,10^{-3})}.
 \label{eq:field-normalization}
\end{equation}
This definition is used for the source field $Q_0$ and for fields compared during distillation. The final ranking field follows~\eqref{eq:spatial-ranking}; the source prefix remains a fixed Gate feature extractor and is included in forward computation.

\subsection{Mask Construction and Differentiable Backward Pass}

At inference, the area head produces $a\in(0.05,0.35)$. The selected pixel count is $k=\operatorname{round}(aHW)$ and is clamped to the valid pixel range. The implementation sorts the flattened $Q$ in descending order and selects indices with zero-based rank below $k$. Equal scores are truncated according to the index order returned by the sorting implementation, so the output always contains exactly $k$ pixels.

Spatial Gate training uses the fixed-budget mask in~\eqref{eq:soft-mask}. For each image, the threshold search interval extends the current field minimum and maximum by $16/\kappa$, and bisection solves for a mask mean of $a_0=0.25$. During the first 80\% of updates, $\kappa$ increases smoothly from 4 to 20 and then remains at 20. Backpropagation uses the implicit gradient in~\eqref{eq:soft-mask-gradient}.

\subsection{Stage-Wise Optimization Settings}
\label{app:training-settings}

Table~\ref{tab:stagewise-training} lists the principal training dependencies and update sets for the current RAF-DB model. The high-resolution teacher, heavy semantic Gate, and source spatial branch provide supervision or initialization and do not enter the deployed classification path.

\begin{table}[t]
 \centering
 \caption{Stage-wise training of STAR-FER.}
 \label{tab:stagewise-training}
 \setlength{\tabcolsep}{2.2pt}
 \begin{tabular*}{\columnwidth}{@{\extracolsep{\fill}}p{0.25\columnwidth}p{0.29\columnwidth}p{0.35\columnwidth}@{}}
 \toprule
 Stage & Updated components & Main supervision \\
 \midrule
 G/F calibration I & Continuous restorer parameters & Eq.~\eqref{eq:calibration-loss} \\
 G/F calibration II & Restorer and FER layer3/4/FC & Eq.~\eqref{eq:calibration-loss} \\
 Spatial Gate learning & Lightweight visual, spatial, and auxiliary branches & Eqs.~\eqref{eq:spatial-task-loss}--\eqref{eq:gate-distillation} \\
 Continuous budget learning & 4,449-parameter area head & Eq.~\eqref{eq:budget-loss} \\
 \bottomrule
 \end{tabular*}
\end{table}

The two G/F calibration phases each run for seven epochs. Spatial Gate learning uses a fixed $25\%$ budget; the current lightweight initialization is selected at epoch 2, and the heavy semantic Gate teacher is selected at epoch 24. The continuous budget head is trained for 24 epochs with AdamW, an initial learning rate of $3\times10^{-4}$, weight decay $10^{-4}$, and gradient-norm clipping at 5. The first epoch uses linear warm-up, followed by cosine decay to $0.1$ of the initial learning rate.

For budget ranking, samples in a shuffled mini-batch are paired with cyclically shifted samples. Only pairs whose teacher budgets differ by at least 0.05 are retained, and a Softplus ranking objective with temperature 0.03 is applied. The moment term matches the batch means and standard deviations of predicted and teacher budgets. The teacher evaluates $\{0,0.05,\ldots,0.35\}$, while both the continuous targets and final predictions are restricted to $0.05$--$0.35$.

\subsection{Data Degradation and Splits}
\label{app:data-protocol}

\subsubsection{Resolution Degradation}
RAF-DB uses the official aligned $100\times100$ RGB faces as references. For each degradation factor $s\in\{2,3,4,5,6,7,8\}$, Pillow bicubic interpolation downsamples the reference to $n_s\times n_s$, where $n_s=\lceil100/s\rceil$, and then resizes it back to $100\times100$ for caching. The model loader converts each image to a $224\times224$ RGB input.

FERPlus uses its $48\times48$ original images as references. The same bicubic procedure downsamples each image to $\lceil48/s\rceil$ and restores it to $48\times48$. The seven native low-resolution side lengths are 24, 16, 12, 10, 8, 7, and 6 pixels, respectively, followed by conversion to a $224\times224$ RGB input.

\subsubsection{FERPlus Label Processing}
FERPlus provides votes from ten annotators per image. Each image is assigned the category with the most votes. Samples labeled Contempt are removed, leaving seven expression classes.

\subsubsection{Training and Test Splits}
RAF-DB follows the official basic-expression split, with 12,271 training and 3,068 test parent images. FERPlus retains its original training and test splits. All cleaned training samples are used for training, and all cleaned test samples are used for evaluation.

%% file: sections/6_references.tex
\bibliographystyle{IEEEtran}
\bibliography{references}